\documentclass[10pt,twocolumn,letterpaper]{article}

\usepackage[pagenumbers]{cvpr} 
\usepackage{amsmath}
\usepackage{amssymb}
\usepackage{wasysym}
\usepackage{xcolor}
\usepackage{color, colortbl}
\definecolor{green}{HTML}{01B468}
\definecolor{green1}{HTML}{00c957}
\definecolor{citecolor}{HTML}{2980b9}
\definecolor{linkcolor}{HTML}{c0392b}
\definecolor{darkviolet}{HTML}{9400D3}
\definecolor{hotpink}{HTML}{FF69B4}
\definecolor{deeppink}{HTML}{FF1493}

\usepackage{amsthm}
\usepackage{booktabs}
\usepackage{adjustbox}
\usepackage{makecell}
\usepackage{multirow}
\usepackage{epsfig}

\usepackage{algorithm}
\usepackage{algorithmic}
\usepackage{newfloat}
\usepackage{listings}

\usepackage[hyphens]{url}  
\usepackage{graphicx} 
\usepackage{natbib}  
\usepackage{caption} 

\makeatletter
  \newcommand\figcaption{\def\@captype{figure}\caption}
  \newcommand\tabcaption{\def\@captype{table}\caption}
\makeatother
\newcommand\blfootnote[1]{%
  \begingroup
  \renewcommand\thefootnote{}\footnote{#1}%
  \addtocounter{footnote}{-1}%
  \endgroup
}

\definecolor{cvprblue}{rgb}{0.21,0.49,0.74}
\usepackage[pagebackref,breaklinks,colorlinks,citecolor=cvprblue]{hyperref}

\def\paperID{} 
\def\confName{}
\def\confYear{}

\title{Point-PEFT: Parameter-Efficient Fine-Tuning for 3D Pre-trained Models}

\author{
Yiwen Tang\textsuperscript{\rm1*}, 
Ray Zhang\textsuperscript{\rm1*},  
Zoey Guo\textsuperscript{\rm1*},\\
Dong Wang\textsuperscript{\rm1},
Zhigang Wang\textsuperscript{\rm1},
Bin Zhao\textsuperscript{\rm1,2},
Xuelong Li\textsuperscript{\rm1,2},
\vspace{0.2cm}\\
\textsuperscript{\rm 1} Shanghai AI Laboratory,
\textsuperscript{\rm 2} Northwestern Polytechnical University\\
}

\begin{document}
\maketitle
\blfootnote{*\ Equal Contribution.}
\begin{abstract}

The popularity of pre-trained large models has revolutionized downstream tasks across diverse fields, such as language, vision, and multi-modality.
To minimize the adaption cost for downstream tasks, many \textbf{P}arameter-\textbf{E}fficient \textbf{F}ine-\textbf{T}uning (\textbf{PEFT}) techniques are proposed for language and 2D image pre-trained models. 
However, the specialized PEFT method for 3D pre-trained models is still under-explored. To this end, we introduce \textbf{Point-PEFT}, a novel framework for adapting point cloud pre-trained models with minimal learnable parameters. Specifically, for a pre-trained 3D model, we freeze most of its parameters, and only tune the newly added PEFT modules on downstream tasks, which consist of a Point-prior Prompt and a Geometry-aware Adapter. The Point-prior Prompt adopts a set of learnable prompt tokens, for which we propose to construct a memory bank with domain-specific knowledge, and utilize a parameter-free attention to enhance the prompt tokens. The Geometry-aware Adapter aims to aggregate point cloud features within spatial neighborhoods to capture fine-grained geometric information through local interactions. Extensive experiments indicate that our Point-PEFT can achieve better performance than the full fine-tuning on various downstream tasks, while using \textbf{only 5\% of the trainable parameters}, demonstrating the efficiency and effectiveness of our approach. Code is released at \url{https://github.com/Ivan-Tang-3D/Point-PEFT}.
\end{abstract}

\section{Introduction}
The recent advancements in large-scale pre-training with numerous data have gained widespread attention in both industry and academia.
In natural language processing, GPT series~\cite{brown2020language, radford2019language} pre-trained by extensive text corpora exhibit superior language generative capabilities and interactivity. 
For 2D image recognition, ViT~\cite{dosovitskiy2020image} and the multi-modal CLIP~\cite{radford2021learning} can also reveal strong visual generalizability and robustness.
However, the full fine-tuning of these large models normally requires substantial computation resources. 
To alleviate this, many efforts on parameter-efficient fine-tuning (PEFT) have been proposed and applied in both language and image domains, significantly reducing the consumption of tuning resources. 
The principal concept involves freezing most trained parameters in large models, and optimizing only the newly inserted PEFT modules on downstream tasks. 
The popular techniques include adapters~\cite{houlsby2019parameter}, prompt tuning~\cite{lester2021power, jia2022visual}, Low-Rank Adaptation (LoRA)~\cite{hu2021lora}, and side tuning~\cite{zhang2020side}.

\begin{figure}[t!]
\vspace{0.6cm}
    \includegraphics[width=\linewidth]{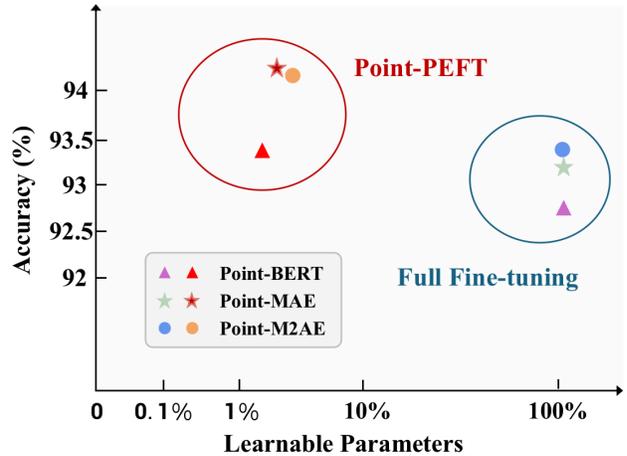}
  \caption{\textbf{Our Point-PEFT vs. Full Fine-tuning} on ModelNet40~\cite{wu20153d} dataset. We compare the fine-tuning of three popular pre-trained models, Point-BERT~\cite{yu2022point}, Point-MAE~\cite{pang2022masked}, and Point-M2AE~\cite{zhang2022point}, where our Point-PEFT achieves superior performance and parameter efficiency.}
  \label{teaser}
\end{figure}

In 3D domains, pre-trained models for point clouds have also shown promising results, e.g., Point-MAE~\cite{pang2022masked}, Point-M2AE~\cite{zhang2022point}, and I2P-MAE~\cite{zhang2023learning}. 
However, the downstream adaption of these 3D transformers is dominated by expensive full fine-tuning, and the specialized 3D PEFT method still remains an open question.
Therefore, inspired by the success in language and 2D methods, we ask the question: \textit{\textbf{can we develop a PEFT framework specialized for 3D point clouds with both efficiency and effectiveness?}}

To tackle this issue, we propose \textbf{Point-PEFT}, a novel parameter-efficient fine-tuning framework for 3D pre-trained models, as shown in Figure~\ref{teaser}. 
Aiming at the sparse and irregular characters of point clouds, we introduce a Point-prior Prompt and a Geometry-aware Adapter, which can efficiently incorporate downstream 3D semantics into the pre-trained models.
On different downstream 3D tasks, we freeze most of the pre-trained parameters, and only fine-tune the task-specific heads and our Point-PEFT components, which we illustrate as follows:

\vspace{0.1cm}
\begin{itemize}
    \item \textbf{Point-prior Prompt.} Before every transformer block, we prepend a set of learnable prompt tokens to the input point cloud tokens, which are enhanced by a proposed point-prior bank with parameter-free attention mechanisms. The bank is constructed by downstream training-set 3D features, which enhances prompt tokens with domain-specific 3D knowledge.
\vspace{0.1cm}
    \item \textbf{Geometry-aware Adapter.} Within each transformer block, we insert the Geometry-aware Adapter after the pre-trained self-attention layer and Feed-Forward Networks (FFN). As the pre-trained attention mainly explores long-range dependencies of the global shape, our adapters are complementary to aggregate local geometric information and grasp the fine-grained 3D structures.
\end{itemize}

\vspace{0.2cm}

With the proposed two components, our Point-PEFT achieves better performance than full fine-tuning, while utilizing only 5\% of the trainable parameters. As an example, for the pre-trained Point-MAE~\cite{pang2022masked}, Point-PEFT with 0.8M parameters attains 94.2\% on ModelNet40~\cite{wu20153d}, and 89.1\% on ScanObjectNN~\cite{uy2019revisiting}, surpassing the full fine-tuning with 22.1M parameters by +1.0\% and +1.0\%, respectively. We also evaluate Point-PEFT on other 3D pre-trained models with competitive results and efficiency, e.g., Point-BERT~\cite{yu2022point} and Point-M2AE~\cite{zhang2022point}, which fully indicates our generalizability and significance.

\vspace{0.1cm}
The contributions of our paper are as follows:

\begin{itemize}
    \item We propose Point-PEFT, a specialized PEFT framework for 3D pre-trained models, which achieves competitive performance to full fine-tuning, and significantly reduces the computational resources.
\vspace{0.1cm}
    \item We develop a Geometry-aware Adapter to extract fine-grained local geometries, and a Point-prior Prompt with parameter-free attention, leveraging domain-specific knowledge to facilitate the downstream fine-tuning.
\vspace{0.1cm}
    \item Extensive experiments indicate the superior effectiveness and efficiency of our approach, which has the potential to serve as a 3D PETT baseline for future research. 
\end{itemize}
\begin{figure*}[t!]
\vspace{-0.1cm}
    \includegraphics[width=\linewidth]{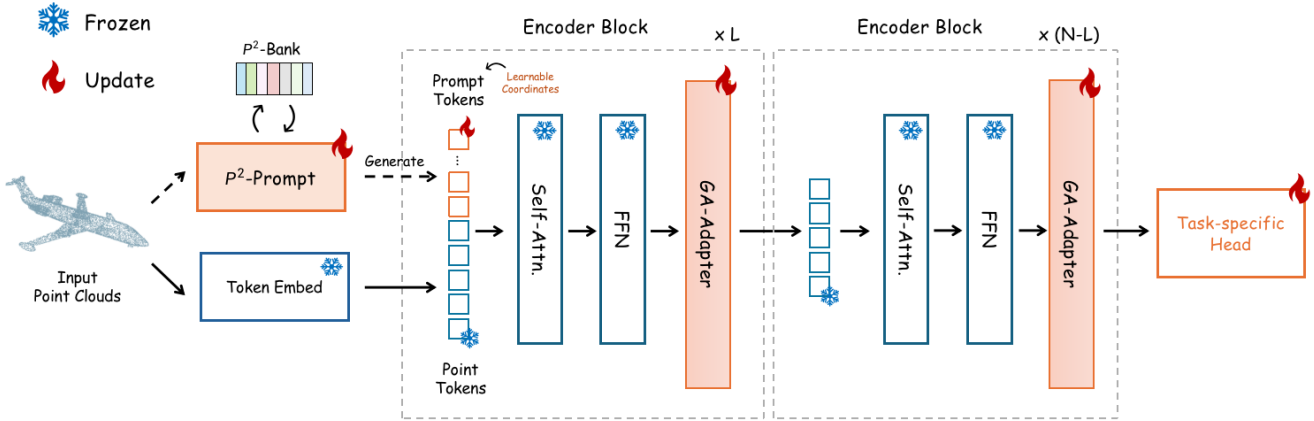}
  \caption{\textbf{Overall Pipeline of Point-PEFT.} 
  For efficiently fine-tuning a pre-trained 3D encoder, our Point-PEFT contains two components: a Point-prior Prompt ($P^2$-Prompt) in the first $L$ blocks, which aggregates prior 3D knowledge from a $P^2$-Bank module, and a Geometry-aware Adapter inserted at the end of each block to effectively grasp the local geometric information.}
  \label{pipeline}
\end{figure*}

\vspace{0.2cm}
\section{Related Work}

\paragraph{Pre-training in 3D Vision.} 
The recent spotlight in the 3D domain has shifted from the supervised training~\cite{qi2017pointnet++,qi2017pointnet} towards self-supervised and large-model pre-training methods due to the challenge of data scarcity. These approaches adopt pretext tasks to pre-train large models, learning the latent representations embedded within point clouds. 
When fine-tuning on downstream tasks~\cite{qu2024livescene,qu2024implicit,duan2024causal}, such as classification~\cite{wu20153d, uy2019revisiting}, segmentation~\cite{yi2016scalable}, and 3D visual grounding~\cite{guo2023viewrefer}, the pre-trained weights are optimized to acquire the knowledge related to the task. 
Some prior works~\cite{afham2022crosspoint, xie2020pointcontrast} utilize contrastive pretext tasks to pre-train the model, discriminating the different views of a single instance from views of other instances. 
Some concurrent works~\cite{zhang2022pointclip, zhu2022pointclip} follow a training-free paradigm, leveraging pre-trained models like CLIP~\cite{radford2021learning} for downstream tasks. 
More research works~\cite{pang2022masked, yu2022point} introduce the masked point modeling strategy for a stronger 3D encoder. 
Point-BERT~\cite{yu2022point} employs a point tokenizer to obtain the point tokens. 
Then the Encoder-Decoder architecture is used to model and predict masked point tokens. 
Point-MAE~\cite{pang2022masked} and Point-M2AE~\cite{zhang2022point} directly utilize Masked Autoencoders (MAE)~\cite{he2022masked}, achieving superior representation capabilities. 
Recently, I2P-MAE~\cite{zhang2023learning}, ACT~\cite{dong2022autoencoders}, and Point-Bind~\cite{point-bind} integrate rich knowledge from pre-trained multi-modal encoders to assist in 3D learning, indicating the potential of introducing external guidance. Joint-MAE~\cite{guo2023joint} and PiMAE~\cite{chen2023pimae} conduct 2D-3D joint pre-training to incorporate cross-modal knowledge. Despite the success of 3D pre-training, the adaption for downstream tasks still demands the resource-intensive full fine-tuning method. Thus, we explore the PEFT techniques in the 3D domain for parameter-efficient fine-tuning.

\paragraph{Parameter-efficient Fine-tuning.}
Given that the full fine-tuning of large models is both computationally intensive and resource-demanding, the Parameter-Efficient Fine-Tuning (PEFT) approaches are proposed to address the challenge by freezing the trained weights and introducing the newly trainable modules.
Various PEFT techniques have been proposed with favorable performance, including adapters~\cite{houlsby2019parameter,gao2021clip}, prompt tuning~\cite{lester2021power, jia2022visual}, Low-Rank Adaptation (LoRA)~\cite{hu2021lora, he2021towards}, bias tuning~\cite{zaken2021bitfit} and side tuning~\cite{zhang2020side, sung2022lst}.
Specifically, the adapter tuning inserts additional bottleneck-shaped neural networks within blocks of the pre-trained model to learn task-specific representations. 
Prompt tuning facilitates task adaption by prepending natural language prompts or learnable prompt tokens to the input. 
The LoRA technique employs a low-rank decomposition approach to learn the adaptation matrix in each block.
Bias tuning achieves competitive results by adjusting the model's bias terms, and side tuning only adjusts side modules concurrent to pre-trained networks. One concurrent work IDPT~\cite{zha2023instance} also introduces PEFT into 3D point cloud transformers, but only investigates prompt tuning methods. In this paper, we propose a PEFT framework specialized for the 3D domain, which introduces a Point-prior Prompt and Geometry-aware Adapter. Different from existing techniques, the former utilizes 3D domain-specific knowledge to enhance the prompt tokens, and the latter grasps the local geometric information.

\section{Method}
We illustrate the details of our Point-PEFT framework for efficiently fine-tuning 3D point cloud pre-trained models. 
We first present our overall pipeline in Section 3.1. Then, in Section 3.2 and 3.3, we respectively elaborate on the designs for Point-prior Prompt and Geometry-aware Adapter.

\subsection{Overall Pipeline}
\label{overall}

As shown in Figure~\ref{pipeline}, given a pre-trained 3D transformer $\operatorname{E_{3D}(\cdot)}$ with 12 blocks and a specific downstream task, we freeze most of its parameters for fine-tuning, and only update our introduced Point-PEFT modules, task-specific heads, and all the bias terms within the transformer blocks~\cite{zaken2021bitfit}.



For an input point cloud $PC$, we follow the original pipeline of the pre-trained transformer to first encode it into $M$ point tokens via the `Token Embed' module, which normally consists of a mini-PointNet~\cite{qi2017pointnet}. We denote the point tokens as $F_0 \in \mathbb{R}^{M \times D}$, where $D$ denotes the feature dimension of the transformer. Then, we prepend our proposed $K$-length Point-prior Prompt, denoted as $P_0\in \mathbb{R}^{K \times D}$ to these point tokens. Each token of $P_0$ is assigned with a learnable 3D coordinate to indicate its spatial location.
Specifically, $P_0$ is generated by a `$P^2\text{-Prompt$(\cdot)$}$' module, which takes the point cloud $PC$ as input, and aggregates domain-specific knowledge from a constructed `$P^2\text{-Bank}$', as shown in Figure~\ref{pipeline}. We formulate it as
\begin{align}
    P_0 &= P^2\text{-Prompt}(PC),\\
    C_0 &= \text{Concat}\left(P_0, F_0\right),
\end{align}
where $C_0\in R^{(K+M)\times D}$ denotes the initial input tokens for the first transformer block.

For the $i$-th transformer block ($2\leq i\leq 12$), we denote the point tokens from the last block as $F_{i-1}$, and concatenate them with the Point-prior Prompt $P_{i}$, which obtains $C_i$ as the input tokens. Then, we feed $C_i$ into the pre-trained self-attention layer and the Feed-Forward Networks (FFN) with residual connections. After that, we adopt our introduced Geometry-aware Adapter (`GA-Adapter') to encode fine-grained local 3D structures, formulated as 
\begin{align}
    C_i' &= \text{FFN}\big(\text{Self-Attn.}(C_i)\big),\\
    F_{i} &= \text{GA-Adapter}(C_i'),
\end{align}
where $F_i$ denotes the output point features from the $i$-th block. 
Note that the prompt tokens are only adopted in the earlier $L$ blocks for better adapting shallower point features.
After all 12 transformer blocks, the learnable downstream task head is adopted to produce the final predictions.

\begin{figure*}[ht!]
\vspace{-0.1cm}
\begin{minipage}[t]{0.58\textwidth}
\raisebox{0.3cm}{\centering{\scalebox{1.02}{\includegraphics[width=\textwidth]{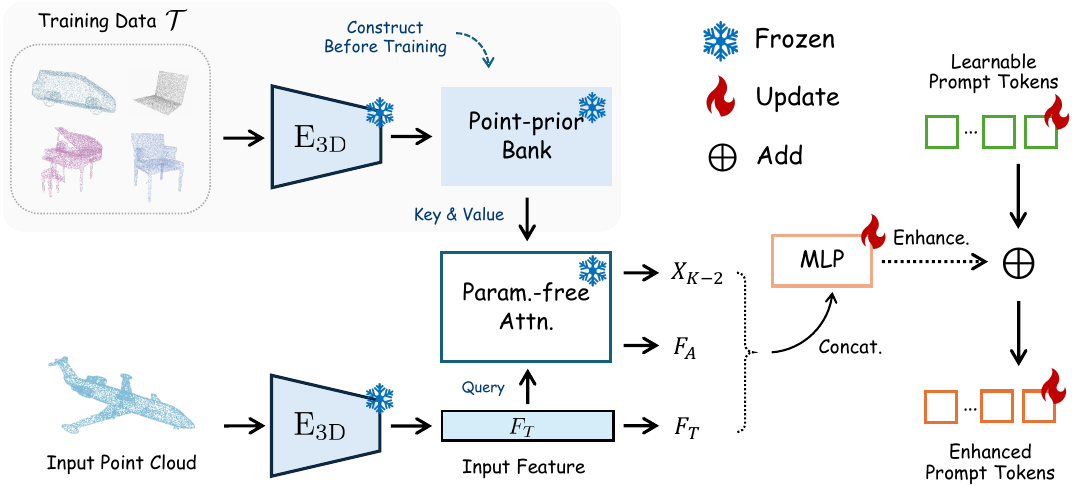}}}}
\figcaption{\textbf{Point-prior Prompt.} To generate the prompt token with 3D prior knowledge, we construct a point-prior bank before fine-tuning, and conduct parameter-free attention for feature aggregation, which adaptively enhances the learnable prompt token with domain-specific semantics.}
\label{prompt_1}
\end{minipage}
\hspace{0.1in}
\begin{minipage}[t]{0.36\textwidth}
\centering{\scalebox{0.55}{\includegraphics[width=\textwidth]{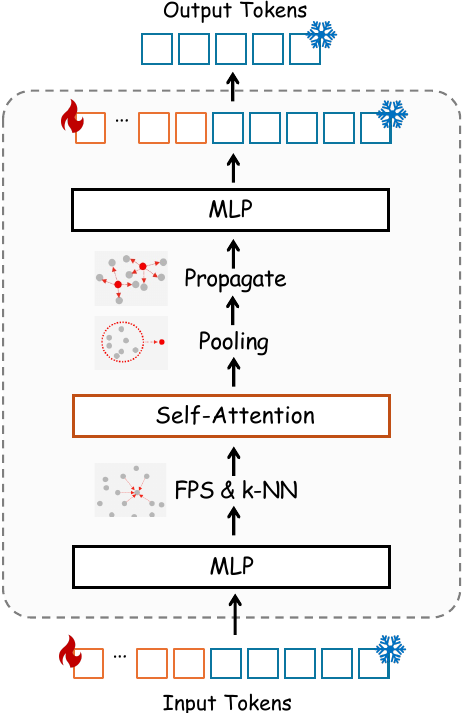}}}

\figcaption{\textbf{Geometry-aware Adapter.} Inserted into every transformer block, the adapter aims to extract the fine-grained geometric information by local interactions.}
\label{Adapter_1}
\end{minipage}
\end{figure*}

\subsection{Point-prior Prompt}
As shown in Figure~\ref{prompt_1}, our adopted prompt tokens $P_i$ for the $i$-th transformer block are generated by a constructed point-prior bank and parameter-free attention.

To create the bank, we employ the pre-trained 3D transformer $\operatorname{E_{3D}}$ to encode all the point clouds in the downstream training dataset $\mathcal{T}$, denoted as $\{PC_n\}_{n=1}^{|\mathcal{T}|}$. Then, we concatenate the training-set features along the sample dimension, and store them as the prior knowledge of the downstream 3D domain, formulated as
\begin{align}
\begin{split}
    X &= \text{Concat}\left(\{E_{3D}(PC_n)\}_{n=1}^{|\mathcal{T}|}\right) \in \mathbb{R}^{|\mathcal{T}| \times D}.
\end{split}
\end{align}
Such point-prior bank is constructed in a training-free manner, referring to existing cache-based methods in 2D~\cite{zhang2022tip,zhang2023prompt,rong2023retrieval} and 3D~\cite{zhang2022nearest,zhang2023parameter,zhu2023less} vision.

For the input point cloud $PC$, we also utilize the pre-trained 3D transformer $\operatorname{E_{3D}}$ to acquire its 3D feature, denoted as $F_T \in \mathbb{R}^{1\times D}$. 
Then, we conduct the parameter-free attention for $F_T$ to adaptively aggregate informative semantics from the point-prior bank $X$.
In detail, the input point cloud feature $F_T$ serves as the query, and the pre-encoded training-set features $X$ of the point-prior bank serve as the key and value. 
Within the attention mechanism, we first calculate the cosine similarity $S$ between the query and key, formulated as
\begin{align}
\begin{split}
    S = \frac{F_TX^{\top}}{|F_T| \cdot |X|} \in \mathbb{R}^{1 \times |\mathcal{T}|}.
\end{split}
\end{align}
The similarity $S$ denotes the attention scores of the input point cloud to all prior training-set 3D knowledge.
Subsequently, we sort the similarity $S$ and obtain the top-$(K-2)$ scores as \( S_{K-2} \in \mathbb{R}^{1 \times (K-2)} \). Accordingly, we select the corresponding $(K-2)$ training-set features in the value, denoted as \(X_{K-2} \in \mathbb{R}^{(K-2)\times D} \).
On top of this, we aggregate the prior knowledge in \(X_{K-2} \in \mathbb{R}^{(K-2)\times D} \) weighted by \( S_{K-2} \in \mathbb{R}^{1 \times (K-2)} \) as 
\begin{align}
\begin{split}
    F_A = \text{Softmax}(S_{K-2})X_{K-2}, 
\end{split}
\end{align}
where $F_A$ represents the input point cloud feature after aggregating the prior knowledge from the point-prior bank.

After that, we concatenate the original feature $F_T$ with $F_A$ and $X_{K-2}$ to obtain a comprehensive representation of the current point cloud and all its relevant 3D prior semantics, which is then transformed by an MLP with bottleneck layers. We formulate it as 
\begin{align}
    P_{prior} &= \text{MLP}\big(\text{Concat}(F_T, F_A, X_{K-2})\big),
\end{align}
where $P_{prior} \in \mathbb{R}^{K\times D}$ denotes the point prompt adaptively generated by the point-prior bank.

For the $i$-th transformer block, we acquire the final Point-prior prompt by element-wisely adding $P_{prior}$ with a set of learnable prompt tokens, $R_i \in \mathbb{R}^{K\times D}$, formulated as
\begin{align}
    P_{i} &= R_i + P_{prior}.
\end{align}
The former component, $R_i$ denotes the learnable downstream knowledge specific to the $i$-th block, while the latter adaptively enhances it by the prior domain-specific semantics, contributing to better fine-tuning performance.

\subsection{Geometry-aware Adapter}
In the $i$-th transformer block, after being processed by the pre-trained self-attention layer and FFN, the point tokens $C'_i \in \mathbb{R}^{(K+M) \times D}$ are fed into the Geometry-aware Adapter.
The adapter aims to grasp the fine-grained geometric information through local aggregation, complementary to the pre-trained global interactions of the self-attention layer.

As shown in Figure~\ref{Adapter_1}, the input $C'_i$ is first transformed by an MLP with bottleneck layers, obtaining $T_{i} \in \mathbb{R}^{(K+M) \times D}$. 
Then, we adopt farthest point sampling (FPS) to downsample the token number from $(K+M)$ to $N$, denoted as $T^c_{i}$, which serves as a set of local centers. After that, we acquire the neighboring points, $T^n_{i}$, for each local center by the $k$-nearest neighbor ($k$-NN) algorithm. We formulated the above process as
\begin{align}
    T^c_{i} &= \text{FPS}(T_i) \in \mathbb{R}^{N \times D},\\
    T^n_{i} &= \text{$k$-NN}(T^c_{i}, T_{i}) \in \mathbb{R}^{N \times k \times D}.
\end{align}
To grasp the fine-grained local semantics within each group, $T_{i}^n$ is fed into a self-attention layer for intra-group interactions. 
The weights of the self-attention layer are shared across all transformer blocks, which effectively reduces the trainable parameters. We formulate it as
\begin{align}
\begin{split}
    {{T_{i}}^{n}}' = \text{Self-Attn.}({T_{i}}^n).
\end{split}
\end{align}
On top of this, we utilize a pooling operation to integrate the features within each local neighborhood, and conduct a weighted summation between the integrated features and the original local-center features as
\begin{align}
\begin{split}
    {{T_i}^c}' = \text{Pooling}({{T_{i}}^{n}}') + \alpha \cdot {T_i}^c,
\end{split}
\end{align}
where ${{T_i}^c}' \in \mathbb{R}^{N \times D}$ denotes the enhanced local-center features with fine-grained geometries, and $\alpha$ denotes a balance factor.
Finally, referring to PointNet++~\cite{qi2017pointnet++}, we propagate ${{T_i}^c}'$ from each local center to its corresponding $k$ neighboring points with a weighted summation as 
\begin{align}
\begin{split}
    {T_i}' = \text{Propagate}({{T_i}^c}') + \beta \cdot T_i,
\end{split}
\end{align}
where $T_{i}' \in \mathbb{R}^{(K+M) \times D}$, and $\beta$ denotes a balance factor. After incorporating the fine-grained 3D semantics, $T_{i}'$ is further processed by an MLP to obtain the output tokens of the $i$-th block, $F_i$.
\begin{figure}[t!]
\centering
\small
\centering
	\begin{tabular}{lc c c}
	\toprule
		\makecell*[l]{\textbf{Method}} &\textbf{\#Param (M)} &\textbf{Acc. (\%)}\\
		\cmidrule(lr){1-1} \cmidrule(lr){2-2} \cmidrule(lr){3-3}
            \color{gray}{\textit{Training from Scratch}}\vspace{0.06cm}\\
            Point-NN &0.0 &64.9\\
	    PointNet &3.5 &68.0\\
	    PointNet++ &1.5 &77.9\\
            PointMLP &14.9  &85.2\\
            Point-PN\textcolor{red}{$^\dagger$} &0.8 &87.1\\
            \midrule
            \color{gray}{\textit{Self-supervised Pre-training}}\vspace{0.06cm}\\
	    Point-BERT &22.1 &83.1\vspace{0.02cm}\\
            \rowcolor{gray!8} \textbf{w/ Point-PEFT}  &\textbf{0.6} &\textbf{85.0}\\
            &\textcolor{green}{\textbf{$\downarrow$97.4\%}}&\textcolor{blue}{\textbf{+1.9}}\vspace{0.02cm}\\
            \cmidrule(lr){1-3}
            
            Point-M2AE  &15.3 &86.4\vspace{0.02cm}\\
            Point-M2AE\textcolor{red}{$^\dagger$} &15.3 &88.1\vspace{0.05cm}\\
            
            \rowcolor{gray!8} \textbf{w/ Point-PEFT}\textcolor{red}{$^\dagger$} &\textbf{0.7} &\textbf{88.2}\\
            &\textcolor{green}{\textbf{$\downarrow$95.4\%}}&\textcolor{blue}{\textbf{+0.1}}\vspace{0.02cm}\\
            \cmidrule(lr){1-3}
            
            Point-MAE &22.1 &85.2\vspace{0.02cm}\\
            Point-MAE\textcolor{red}{$^\dagger$} &22.1 &88.1\vspace{0.02cm}\\
            
            \rowcolor{gray!8}  \textbf{w/ Point-PEFT}\textcolor{red}{$^\dagger$}  &\textbf{0.7} &\textbf{89.1}\\
            &\textcolor{green}{\textbf{$\downarrow$96.8\%}}&\textcolor{blue}{\textbf{+1.0}}\vspace{0.02cm}\\
	  \bottomrule
	\end{tabular}
\tabcaption{\textbf{Real-world 3D Classification on ScanObjectNN}. We report the number of learnable parameters (\#Param) and the accuracy (\%) on the "PB-T50-RS" split of ScanObjectNN. 
\textcolor{red}{$^\dagger$} indicates utilizing a stronger data augmentation~\cite{zhang2023learning} during fine-tuning.}
 \label{scan_cls}
\end{figure}

\begin{figure}[t!]
\centering
\small
\centering
	\scalebox{0.97}{\begin{tabular}{lc c c}
	\toprule
		\makecell*[l]{\textbf{Method}} &\textbf{\#Param (M)}  &\textbf{Acc. (\%)}\\
		\cmidrule(lr){1-1} \cmidrule(lr){2-2} \cmidrule(lr){3-3}
            \color{gray}{\textit{Training from Scratch}}\vspace{0.06cm}\\
            Point-NN &0.0 &81.8\\
	    PointNet &\centering 3.5  &89.2\\
	    PointNet++ &\centering 1.5  &90.7\\
            PointCNN &\centering 0.6   &92.2\\
	    DGCNN  &\centering 1.8  &92.9\\
            PCT &\centering 2.9  &93.2\\
            Point-PN &0.8 &93.8\\
            PointMLP &\centering 14.9  &94.1\\
	                \midrule
           \color{gray}{\textit{Self-supervised Pre-training}}\vspace{0.06cm}\\
	    Point-BERT &\centering 22.1  &92.7\vspace{0.02cm}\\
            \rowcolor{gray!8} \textbf{w/ Point-PEFT}  &\centering\textbf{0.6}  
            &\textbf{93.4}\\
            &\textcolor{green}{\textbf{$\downarrow$97.4\%}}&\textcolor{blue}{\textbf{+0.7}}\vspace{0.02cm}\\
            \cmidrule(lr){1-3}
            Point-M2AE  &\centering 15.3  &93.4\vspace{0.02cm}\\
            \rowcolor{gray!8} \textbf{w/ Point-PEFT} &\centering\textbf{0.6}  &\textbf{94.1}\\
            &\textcolor{green}{\textbf{$\downarrow$96.1\%}}&\textcolor{blue}{\textbf{+0.7}}\vspace{0.02cm}\\
            \cmidrule(lr){1-3}
            Point-MAE &\centering 22.1  &93.2\vspace{0.02cm}\\
            \rowcolor{gray!8} \textbf{w/ Point-PEFT}  &\centering\textbf{0.8}  &\textbf{94.2}\\
            &\textcolor{green}{\textbf{$\downarrow$96.4\%}}&\textcolor{blue}{\textbf{+1.0}}\vspace{0.02cm}\\
	  \bottomrule
	\end{tabular}}
\tabcaption{\textbf{Synthetic 3D Classification on ModelNet40}. We report the number of learnable parameters (\#Param) and the accuracy (\%) on ModelNet40. Note that, during the testing process, we do not employ the voting strategy.}
 \label{modelnet_cls}
\end{figure}

\section{Experiments}
We evaluate the performance of our proposed Point-PEFT framework for 3D shape classification. 
We utilize three pre-trained models (Point-BERT~\cite{yu2022point}, Point-MAE~\cite{pang2022masked}, and Point-M2AE~\cite{zhang2022point}) as our baselines. 
Please refer to the Supplementary Material for experiments of part segmentation and additional ablation studies.
\subsection{Experimental Settings}
\subsubsection{ScanObjectNN.}
The ScanObjectNN~\cite{uy2019revisiting} dataset is a real-world 3D point cloud classification dataset, containing about 15,000 3D objects from 15 distinct categories. We focus on the hardest "PB-T50-RS" split, where the rotation (R) and scaling (S) augmentation methods are applied to objects.
For all considered models, we employ the AdamW optimizer~\cite{loshchilov2017decoupled} coupled with a cosine learning rate decay strategy. 
The initial learning rate is set as \(0.0005\), with a weight decay factor of \(0.05\). 
We fine-tune the models in 300 epochs, utilizing a batch size of 32.
As shown in Table~\ref{scan_cls}, \textcolor{red}{\( \dagger \)} indicates that the fine-tuning utilizes a stronger data augmentation in I2P-MAE~\cite{zhang2023learning}, including random scaling, translation, and rotation. Otherwise, we only adopt random scaling and translation.
Respectively for Point-BERT, Point-MAE, and Point-M2AE, we set the prompting layers and prompt length ($L$, $K$) as (6, 5), (6, 10), and (15, 16).

\subsubsection{ModelNet40.}
The ModelNet40 dataset~\cite{wu20153d} comprises a total of 12,311 3D CAD models across 40 categories. 
The point cloud objects are complete and uniform.
For experiments on ModelNet40, we adopt the same fine-tuning settings as ScanObjectNN.
For all models, we adopt the default data augmentation  random scaling and translation. 
Respectively for Point-BERT, Point-MAE, and Point-M2AE, we set the prompting layers and prompt length ($L$, $K$) as (9, 16), (12, 16), and (15, 16).
Note that, during the testing process, we do not employ the voting strategy.

\begin{figure}[t!]
\vspace{-0.1cm}
\centering
\small
\centering
	\scalebox{1.1}{\begin{tabular}{lc c c c}
	\toprule
		\makecell*[l]{Method} &\#Param (M) &Acc. (\%)\\
		\cmidrule(lr){1-1} \cmidrule(lr){2-2} \cmidrule(lr){3-3} 
	    Full Fine-Tuning &\centering 22.1 M &88.1\\
	   \cmidrule(lr){1-3}
	    Prompt Tuning &\centering 0.3 M &83.6\\
            Adapter Tuning &\centering 0.4 M &86.7\\
	    LoRA  &\centering 0.4 M &86.3\\
            Bias Tuning &\centering 0.3 M &85.0\\
	    \rowcolor{gray!8} \textbf{Point-PEFT}  &\centering\textbf{0.7 M} &\textbf{89.1}\vspace{0.02cm}\\
	  \bottomrule
	\end{tabular}}
\tabcaption{\textbf{Ablation Study on Different PEFT Methods}.}
 \label{abl_peft}
\end{figure}

\begin{figure}[t!]
\centering
\small
\centering
	\scalebox{0.95}{\begin{tabular}{c c c|c}
	\toprule
		\makecell*[c]{Point-prior Prompt} &\makecell*[c]{GA-Adapter} &\makecell*[c]{Bias Tuning} &\makecell*[c]{Acc. (\%)}\\
		 \cmidrule(lr){1-1} \cmidrule(lr){2-2} \cmidrule(lr){3-3} \cmidrule(lr){4-4} 
        \rowcolor{gray!6}-&-&-&88.1\\
        \checkmark&-&-&84.7\\
        -&\checkmark&-&87.6\\
        \checkmark&\checkmark&-&88.6\\
        \rowcolor{gray!9} \checkmark&\checkmark&\checkmark& \textbf{89.1}\\
        
	\bottomrule
	\end{tabular}}

\tabcaption{\textbf{Ablation Study on Main Components.}}
\label{component}
\end{figure} 

\begin{figure}[t!]
\centering
\small
\centering
	\begin{tabular}{c c |c}
	\toprule
		\makecell*[c]{Point-prior Bank} &\makecell*[c]{Learnable Positions} &\makecell*[c]{Acc. (\%)}\\
		 \cmidrule(lr){1-1} \cmidrule(lr){2-2} \cmidrule(lr){3-3} 
        \rowcolor{gray!6}-&-&87.9\\
        \checkmark&-&88.4\\
        \rowcolor{gray!9} \checkmark&\checkmark& \textbf{89.1}\\
        
	\bottomrule
	\end{tabular}
\tabcaption{\textbf{Ablation Study on Point-prior Prompt.}}
\label{prompt_component}
\end{figure}

\subsection{Quantitative Analysis}
\subsubsection{Performance on ScanObjectNN.}
As shown in Table~\ref{scan_cls}, our Point-PEFT framework surpasses the full fine-tuning method with less than 5\% trainable parameters. 
The improvements brought by our framework are +1.9\%, +0.1\%, and +1.0\% for Point-BERT, Point-M2AE\textcolor{red}{$^\dagger$}, and Point-MAE\textcolor{red}{$^\dagger$} respectively, indicating our great advantages under complex 3D scenes by the extracted fine-grained geometric information.
Compared to the full fine-tuning method, our Point-PEFT framework has the stronger ability to be adapted to the tasks related to the real-world scanned objects by pre-trained prior knowledge.
\begin{figure}[t!]
\centering
\small
\centering
	\begin{tabular}{c c c|c}
	\toprule
		\makecell*[c]{Local Aggregation} &\makecell*[c]{Self-Attn.} &\makecell*[c]{MLP$_{final}$} &\makecell*[c]{Acc. (\%)}\\
		 \cmidrule(lr){1-1} \cmidrule(lr){2-2} \cmidrule(lr){3-3} \cmidrule(lr){4-4} 
        \rowcolor{gray!6}-&-&-&87.0\\
        \checkmark&-&-&88.2\\
        \checkmark&\checkmark&-&88.7\\
        \rowcolor{gray!9} \checkmark&\checkmark&\checkmark& \textbf{89.1}\\
	\bottomrule
	\end{tabular}
\tabcaption{\textbf{Ablation Study on Geometry-aware Adapter.} {`Local Aggregation' refers to the FPS, $k$-NN, pooling, and propagation operations in the adapter.}}
\label{adapter_component}
\end{figure} 
\begin{figure}[t!]
\vspace{-0.1cm}
    \centering{\scalebox{1}{\includegraphics[width=0.95\linewidth]{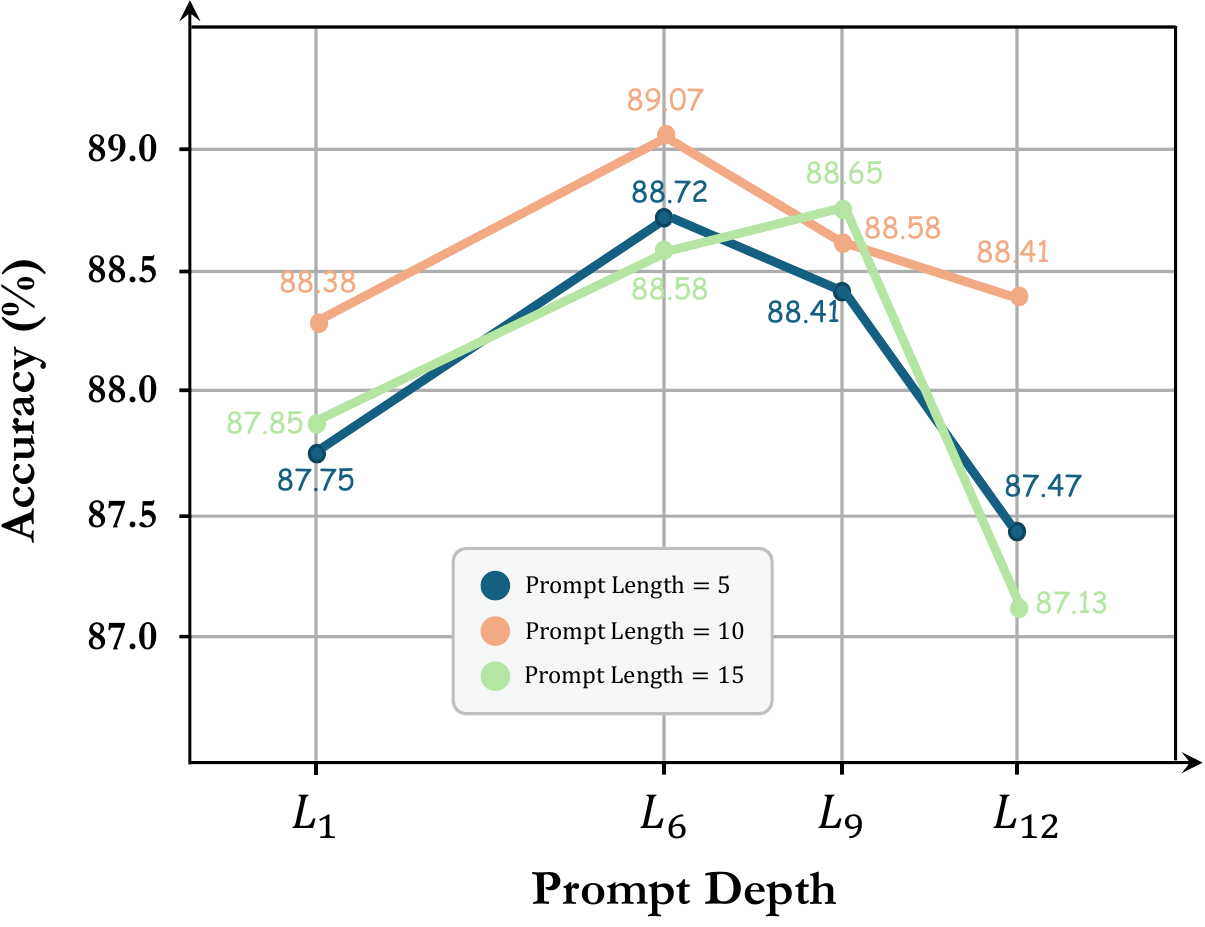}}}
    
  \caption{\textbf{Ablation Study on Prompt Length and Depth.} The deep blue, light orange, and light green lines represent a prompt length of 5, 10, and 15, respectively.}
  \label{depth}
\end{figure}

\subsubsection{Performance on ModelNet40.}
As shown in Table~\ref{modelnet_cls}, employing our Point-PEFT framework with less than 4\% learnable parameters, we achieve performances of 93.4\%, 94.1\%, and 94.2\% for Point-BERT, Point-M2AE, and Point-MAE respectively with the gains of +0.7\%, +0.7\%, and +1.0\%.
These results point out the effectiveness of our framework in handling the sparse and irregular point cloud features.
For the synthetic point cloud objects, the Point-PEFT framework could grasp the global shape and understand the local 3D structures concurrently.
\section{Ablation Study}

In this section, we conduct extensive ablation studies to explore the effectiveness of different components within our Point-PEFT framework. We adopt Point-MAE\textcolor{red}{$^\dagger$} as the pre-trained model, and report the classification accuracy (\%) on the "PB-T50-RS" split of the ScanObjectNN dataset.

\subsection{Comparison to Traditional PEFT Methods.}
As shown in Table~\ref{abl_peft}, our Point-PEFT framework can surpass conventional PEFT techniques with huge gains, e.g., +5.5\% over Prompt Tuning, +2.4\% over Adapters, +2.8\% over Low-Rank Adaptation (LoRA), and +4.1\% over Bias Tuning. 
The comprehensive experiments have exhibited the superiority of our framework over the traditional PEFT methods, indicating that our proposed method effectively integrates 3D domain-specific knowledge into the PEFT framework.
In contrast to the PEFT techniques in language and 2D image domains, our framework focuses more on the complex and irregular point cloud structures, specifically designed for 3D.

\begin{figure}[t!]
\vspace{-0.1cm}
    \includegraphics[width=\linewidth]{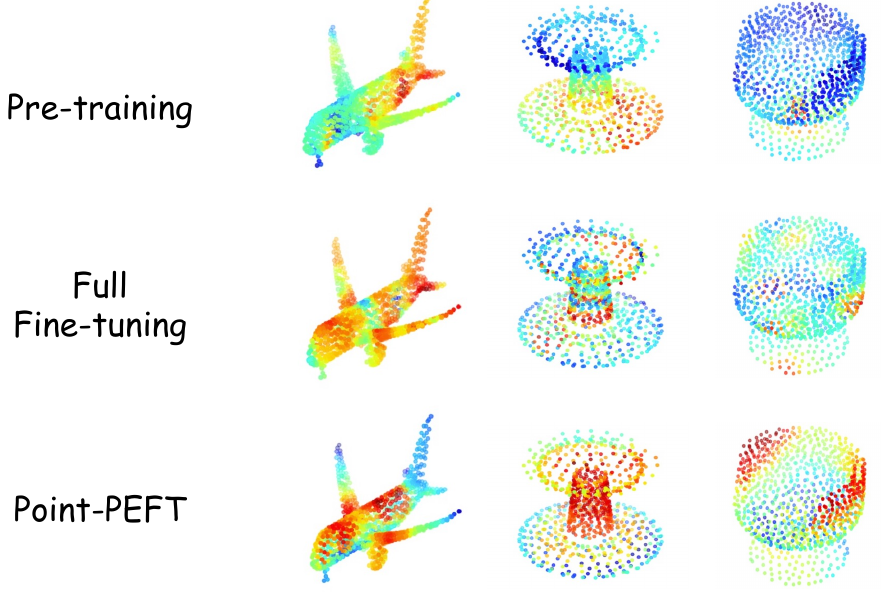}
  \caption{\textbf{Visualization of the Captured Fine-grained Information.} We visualize the point feature responses respectively for the pre-trained model, the full fine-tuning method, and our proposed Point-PEFT. \textbf{The red color indicates higher responses.}}
  \label{visual1}
\vspace{-0.2cm}
\end{figure}

\subsection{Effectiveness of Main Components.}
As shown in Table~\ref{component}, to substantiate the effectiveness of each component, we conduct the ablation experiments by incrementally introducing components to the baseline (Point-MAE\textcolor{red}{$^\dagger$} model) until the complete Point-PEFT framework. 
The light gray row indicates the baseline with the transformer encoder, and the dark gray row represents the complete structure of the Point-PEFT framework.
Introducing either the Point-prior Prompt or the Geometry-aware Adapter component leads to a certain degree of performance degradation compared to full fine-tuning. 
When both components are utilized, the performance has been improved to 88.6\% with a gain of +0.5\%. Further adding Bias Tuning can result in an additional +0.5\% improvement.
Therefore, the experiments indicated the effectiveness of each design within the Point-PEFT framework.

\subsection{Effects of Prompt Length and Depth.}
In Figure~\ref{depth}, we ablate the number of earlier layers applying the prompt tokens (`Prompt Depth') and the `Prompt Length'.
As shown, longer prompt tokens don't necessarily lead to better performance.
For different prompt depths, the length of 10 almost yields the best results, indicating that a moderate length is the most appropriate for 3D prompt learning.
In addition, the optimal depth varies with different prompt token lengths.
Notably, the insertion of prompt tokens in all blocks fails to bring significant performance improvement.
Introducing them in a certain number of earlier blocks is the preferable strategy, suggesting that prompt tokens in earlier layers carry more significance than those in later ones.
\begin{figure}[t!]
\vspace{-0.1cm}
    \includegraphics[width=\linewidth]{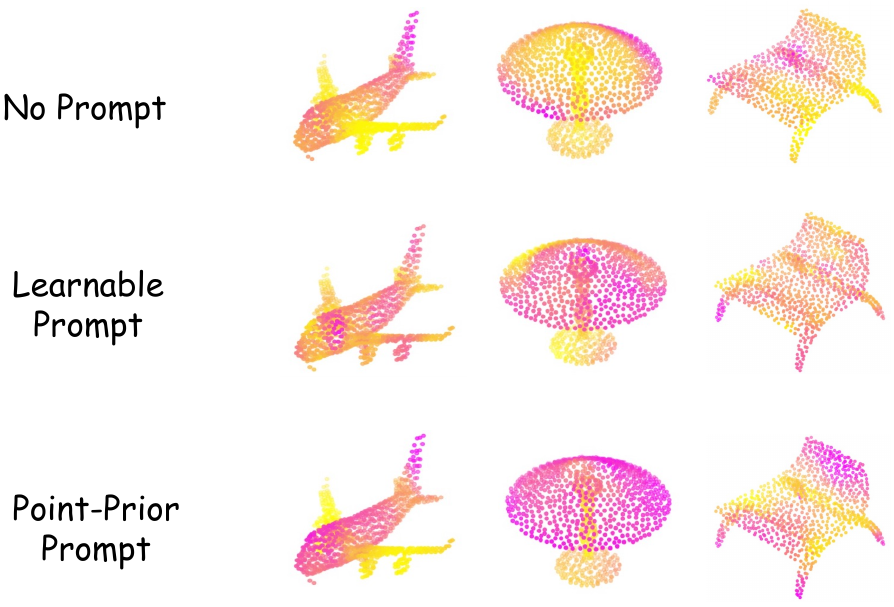}
  \caption{\textbf{Visualization of Different Prompt Tuning Methods.} For the three rows, we respectively visualize the attention scores of the \textbf{[CLS]} token, the learnable prompt token, and the Point-prior prompt token to other point cloud tokens. \textbf{The pink color indicates higher values.}}
  \label{prompttokens}
\vspace{-0.2cm}
\end{figure}

\subsection{Components of Point-prior Prompt.}
As shown in Table~\ref{prompt_component}, we investigate the effects of the point-prior bank and the learnable coordinates for the Point-prior Prompt. The first row in light gray represents that we only utilize the learnable prompt tokens, which are randomly initialized before training. The last row in dark gray shows the complete structure of the Point-prior Prompt.
By constructing the point-prior bank with the 3D domain-specific semantics to enhance the prompt tokens, our method achieves a performance gain of +0.5\%, indicating the importance of prior knowledge enhancement. The assignment of the shared learnable coordinates boosts +0.7\% in performance, which represents the spatial locations of the prompt tokens. The coordinates enable the prompt tokens to engage in the local interactions of the Geometry-aware Adapter alongside the features.
The experiments confirm the effectiveness of each component in the Point-prior Prompt to leverage the prior 3D semantics for downstream tasks.
\subsection{Components of Geometry-aware Adapter.}
In Table~\ref{adapter_component}, we conduct ablation studies by incrementally introducing components of our Geometry-aware Adapter. The row in light gray signifies the baseline adapter, consisting of only an MLP with bottleneck layers. The row in dark gray indicates the complete structure of the Geometry-aware Adapter.
By integrating the local-aggregation operations, comprised of FPS, $k$-NN, pooling, and propagation operations, our approach achieves a significant performance improvement of +1.2\%. These operations effectively extract the 3D fine-grained structures of local neighborhoods. The self-attention layer brings an additional performance gain of +0.5\%, boosting the perception of fine-grained geometric information through the intra-group feature interactions. Lastly, by introducing the final bottleneck-layer MLP to further process point cloud features, the performance is improved by +0.4\%.
The experiments fully demonstrate the effectiveness of each component in our Geometry-aware Adapter to aggregate the local geometric information, which is complementary to the global attention in pre-trained 3D models.

\section{Visualization}

\subsection{Fine-grained Geometric Information.}
The Geometry-aware Adapter captures local geometric knowledge through the interactions within local regions. 
In Figure~\ref{visual1}, we visualize the feature responses based on Point-MAE. In each row, we show the responses for the pre-trained model, the full fine-tuning method and our Point-PEFT respectively, where warm colors indicate the high responses. As shown, compared with others' randomly focusing on the less significant parts or concentrating consistently on the whole object, our Point-PEFT more focuses on the discriminative parts of the objects, such as the wings and engines of airplanes, the brackets and shades of lamps, 
which are critical for distinguishing similar 3D shapes. 
This indicates that our Point-PEFT with the Geometry-aware Adapter not only emphasizes global information but also boosts the understanding of the fine-grained crucial structures. 

\subsection{Different Prompt Tuning Methods.}
The Point-prior Prompt utilizes 3D domain-specific knowledge from the point-prior bank to enhance the prompt tokens. 
In Figure~\ref{prompttokens}, we respectively visualize the attention scores of the [CLS] token, the learnable prompt token,
and the Point-prior prompt token to other point cloud tokens, where the pink color indicates higher values.
As illustrated, the [CLS] token grasps useless information, and the learnable prompt tokens fail to capture the crucial 3D semantics. 
Compared with them, our proposed Point-prior Prompt tokens focus more on the salient and important object parts, such as the fuselages and tails of airplanes, the entire shades of lamps, and the backrests and legs of chairs, which indicates that the Point-prior Prompt with prior pre-trained semantics effectively grasps the critical information and further benefits 3D point cloud understanding.

\section{Conclusion}
In this paper, we introduce Point-PEFT, a parameter-efficient fine-tuning framework specialized for pre-trained 3D models. Our approach achieves comparable performance to full fine-tuning on downstream tasks, while significantly reducing the number of learnable parameters. 
The framework consists of a Geometry-aware Adapter and a Point-prior Prompt. 
The Geometry-aware Adapter leverages local interactions to extract fine-grained geometric information. 
The Point-prior Prompt utilizes pre-trained semantic information to enhance the prompt tokens.
Extensive experiments validate the effectiveness of Point-PEFT. 
We expect Point-PEFT can serve as a baseline for future 3D PEFT research.

\appendix

\section{Implementation Details}

The inner dimension of the bottleneck-shaped MLP in the Point-prior Prompt is set as 8 for all pre-trained models on two datasets. The inner dimension of the first bottleneck-shaped MLP in the Geometry-aware Adapter is decided as 16 for all pre-tarined mdoels.
We also set different N and k for the k-NN algorithm at different pre-trained models. The N and k are set as (32, 16), (32, 8), and (64, 16) for Point-BERT~\cite{yu2022point}, Point-M2AE~\cite{zhang2022point}, and Point-MAE~\cite{pang2022masked} respectively.
On the ScanObjectNN~\cite{uy2019revisiting} and ModelNet40~\cite{wu20153d}, we set the inner dimension of the last bottleneck-shaped MLP in the Adapter as (8, 16, 16) and (16,16, 16) for Point-BERT, Point-M2AE, and Point-MAE respectively.
\begin{table}[t!]
\centering
\small
	\begin{tabular}{lc c}
	\toprule
		 Methods &\#Param(M) &Acc.(\%) \\
		\cmidrule(lr){1-1} \cmidrule(lr){2-2} \cmidrule(lr){3-3} 
            \color{gray}{\textit{Training from Scratch}}\vspace{0.06cm} \\
            Point-NN &0.0 &64.9\\
	    PointNet &3.5 &68.0\\
	    PointNet++ &1.5 &77.9\\
            PointMLP &14.9  &85.2\\
            Point-PN\textcolor{red}{$^\dagger$} &0.8 &87.1\\
            \midrule
            \color{gray}{\textit{Self-supervised Pre-training}}\vspace{0.06cm}\\
	    Point-BERT &22.1 &83.1\vspace{0.02cm}\\
            \rowcolor{gray!8} \textbf{w/ Point-PEFT}  &\textbf{0.6} &\textbf{85.0} \\
            &\textcolor{green}{\textbf{$\downarrow$97.4\%}}&\textcolor{blue}{\textbf{+1.9}}\vspace{0.02cm}\\
            \cmidrule(lr){1-3}
            
            Point-M2AE  &15.3 &86.4\vspace{0.02cm}\\
            \rowcolor{gray!8} \textbf{w/ Point-PEFT} &\textbf{0.7} &\textbf{86.4}\\
            &\textcolor{green}{\textbf{$\downarrow$95.4\%}}&\textcolor{blue}{\textbf{+0.0}}\vspace{0.02cm}\\
            Point-M2AE\textcolor{red}{$^\dagger$} &15.3 &88.1\vspace{0.05cm}\\
            
            \rowcolor{gray!8} \textbf{w/ Point-PEFT}\textcolor{red}{$^\dagger$} &\textbf{0.7} &\textbf{88.2}\\
            &\textcolor{green}{\textbf{$\downarrow$95.4\%}}&\textcolor{blue}{\textbf{+0.1}}\vspace{0.02cm}\\
            \cmidrule(lr){1-3}
            
            Point-MAE &22.1 &85.2\vspace{0.02cm}\\
            \rowcolor{gray!8} \textbf{w/ Point-PEFT} &\textbf{0.7} &\textbf{85.5}\\
            &\textcolor{green}{\textbf{$\downarrow$96.8\%}}&\textcolor{blue}{\textbf{+0.3}}\vspace{0.02cm} \\
            Point-MAE\textcolor{red}{$^\dagger$} &22.1 &88.1\vspace{0.02cm}\\
            
            \rowcolor{gray!8}  \textbf{w/ Point-PEFT}\textcolor{red}{$^\dagger$}  &\textbf{0.7} &\textbf{89.1}\\
            &\textcolor{green}{\textbf{$\downarrow$96.8\%}}&\textcolor{blue}{\textbf{+1.0}}\vspace{0.02cm}\\
	  \bottomrule
	\end{tabular}
\tabcaption{\textbf{Performance of Shape Classification on ScanObjectNN Dataset.}  We report the number of learnable parameters (\#Param) and the accuracy (\%) on the "PB-T50-RS" split of ScanObjectNN~\cite{uy2019revisiting}. \textcolor{red}{$\dagger$} indicates utilizing a stronger data augmentation~\cite{zhang2023learning} during fine-tuning.}
\label{shape_all}
\end{table}

\section{Additional Experiments}
\paragraph{Shape Classification.}
Due to the page limitation, we only provide part of the results of Point-PEFT for 3D shape classification in the main paper. In Table~\ref{shape_all}, we report more quantitative results of Point-PEFT with the stronger data augmentation (random scaling, translation and rotation) and default augmentation method (random scaling and translation), which are denoted as with and without \textcolor{red}{$\dagger$} respectively.
As reported, under both of the two data augmentation settings, our Point-PEFT outperforms the full fine-tuning method based on almost all pre-trained models. This indicates the robustness of Point-PEFT on different pre-trained 3D point cloud models and data augmentation methods.



\begin{figure}[t!]
\small
\centering
	\scalebox{0.9}{\begin{tabular}{lcccc}
	\toprule
		Method &\#Param(M) &mIoU$_C$(\%) &mIoU$_I$(\%)\\
		\cmidrule(lr){1-1} \cmidrule(lr){2-2} \cmidrule(lr){3-3} \cmidrule(lr){4-4} 
            \color{gray}{\textit{Training from Scratch}}\vspace{0.06cm} \\
	    PointNet  &\centering{-}&80.39 &83.7 \\
	    PointNet++  &\centering{-}&81.85 &85.1 \\
	    DGCNN  &\centering{-}&82.33 &85.2 \\
	    \cmidrule(lr){1-4}
            \color{gray}{\textit{Self-supervised Pre-training}}\vspace{0.06cm} \\
            Transformer &27.09 &83.42 &85.1 \\
            MaskPoint &- &84.60 &86.0 \\
	    Point-MAE\textcolor{blue}{$^\ddagger$}  &22.26 &84.19 &86.1 \\
            \rowcolor{gray!8} \textbf{w/ Point-PEFT}  &\textbf{0.45}&83.81&85.3\\
            \cmidrule(lr){1-4}
            Point-M2AE\textcolor{blue}{$^\ddagger$} &13.03&\textbf{84.86}&\textbf{86.5}\\
            \rowcolor{gray!8} \textbf{w/ Point-PEFT} &\textbf{0.39}&84.35&86.1\\
	  \bottomrule
	\end{tabular}}
\tabcaption{\textbf{Part Segmentation on ShapeNetPart~\cite{yi2016scalable}}. `mIoU$_C$' (\%) and `mIoU$_I$' (\%) denote the mean IoU across all part categories and all instances in the dataset, respectively. \textcolor{blue}{$^\ddagger$} denotes that we do not include classification heads into the parameter calculation.}
\label{segment}
\end{figure}

\paragraph{Part Segmentation.}
To evaluate the understanding capacity of Point-PEFT on dense 3D data, we further conduct part segmentation experiments on ShapeNetPart~\cite{yi2016scalable} that contains 14,007 and 2,874 samples for training and validation, respectively. We follow previous works to adopt the same segmentation head and sample 2048 points for each object as input. 
For each pre-trained model, we fine-tune Point-PEFT for 300 epochs with a batch size of 16, and adopt the learning rate as 2$\times$10$^{-4}$. In Table~\ref{segment}, we report the mean IoU across all part categories and instances, denoted as mIoU$_C$ and mIoU$_I$, respectively. As reported, on each pre-trained model, our Point-PEFT attains the competitive performance on both metrics, which illustrates the strong understanding capacity of our designs on the fine-grained dense data. Compared to the full fine-tuning method, the number of learnable parameters has dropped \textbf{98\%} and \textbf{97\%} for the Point-MAE and Point-M2AE respectively.

\begin{table}[t!]
\centering
\small
    \scalebox{0.9}{\begin{tabular}{cccccc}
    \toprule
    \multicolumn{4}{c}{\ \ \ \ \ \ \ \ \ Positions of GA-Adapter.\ \ \ \ \ \ \ \ \ } &\makecell*[c]{\multirow{2}*{\shortstack{\vspace*{2.2pt}\\Insertion\\\vspace*{0.3pt}\\Depth}}} &\makecell*[c]{\multirow{2}*{Acc.(\%)}} \\
    \cmidrule(lr){1-4}
      \ \ \ \ After\ \ \ \  &\ \ \ Before\ \ \  &\ \ Both\ \ \ &\ \ Parallel\ \ \  & &\\
     \cmidrule(lr){1-1}  \cmidrule(lr){2-2}  \cmidrule(lr){3-3}  \cmidrule(lr){4-4}  \cmidrule(lr){5-5} \cmidrule(lr){6-6}  
     \checkmark &- &- &- &6 &88.0\% \\
     \rowcolor{gray!7}\checkmark &- &- &- &12 &\textbf{89.1}\% \\
     - &\checkmark &- &- &6 &88.3\% \\
     - &\checkmark &- &- &12 &88.0\% \\
     - &- &\checkmark &- &6 &86.0\% \\
     - &- &\checkmark &- &12 &86.1\% \\
     - &- &- &\checkmark &6 &87.9\% \\
     - &- &- &\checkmark &12 &88.4\% \\
    \bottomrule
\end{tabular}}
\tabcaption{\textbf{Ablation Study on Adapter Positions and Depth.} The `After', `Before', `Both', 'Parallel' denote block structure of putting the "Geometry-aware Adapter" after, before, wrapping around, and in parallel to the "FFN" layer respectively. The insertion depth denotes the certain number of earlier blocks.}
\label{adapter_pos}
\vspace{-0.4cm}
\end{table}

\section{Additional Ablation Study}
\paragraph{The Adapter Positions and Depth}
We further conduct ablation study on the positions and depth of the proposed Geometry-aware Adapter in Table~\ref{adapter_pos}, where we utilize the final version of Point-PEFT as the baseline and compare the accuracy(\%) on the "PB-T50-RS" split of the ScanObjectNN with Point-MAE\textcolor{red}{$^\dagger$} as the pre-trained model. As shown in Table~\ref{adapter_pos}, the last position within the block and the depth of 12 perform the best. That is because, compared to other positions, after the global interactions among point cloud features through pre-trained self-attention layer and Feed-forward Networks(FFN), the local interactions within the adapter could further boost the extraction of the fine-grained geometric information, which facilitates the fine-tuning process. And the deeper insertion could utilize the high-level knowledge to grasp the better understanding of 3D structures. Note that, for all pre-trained point cloud models, we insert the Geometry-aware Adapter in the last position across all blocks.

{
    \small
    \bibliographystyle{ieeenat_fullname}
    \bibliography{main}
}


\end{document}